\documentclass[lettersize,journal]{IEEEtran}
\usepackage{amsmath,amsfonts}
\usepackage{algorithm}
\usepackage{algpseudocode}
\usepackage{array}
\usepackage[caption=false,font=normalsize,labelfont=sf,textfont=sf]{subfig}
\usepackage{textcomp}
\usepackage{stfloats}
\usepackage{url}
\usepackage{verbatim}
\usepackage{graphicx}
\usepackage{cite}
\usepackage{booktabs}
\usepackage{hyperref}

\title{Trajectory Learning with Graph Representations for~Social~Robot~Navigation}

\author{Berke Kartal$^1$, Burcu Kilic$^1$, Yigit Yildirim$^2$, Emre Ugur$^1$
\thanks{This work has been supported by the \textit{INVERSE} project (no. 101136067), funded by the EU.}
\thanks{$^{1}$Bogazici University $^{2}$CREATE Consortium}
}

\begin{document}

\maketitle

\begin{abstract}

Autonomous mobile robots are expected to exhibit socially compliant navigation for minimizing pedestrian disturbance. While capturing social interactions and incorporating pedestrian motion estimations into decision-making are beneficial for compliance, prior methods fail to address both spatial and temporal characteristics present in real-world data. Reinforcement Learning offers high capability, but it requires hand-crafted reward functions that reduce social behavior to static criteria, limiting its ability to reproduce patterns that exist in real pedestrian behavior. Imitation Learning offers direct training from real-world data but lacks modeling of social interactions and suffers from error accumulation. To this end, we propose an imitation learning framework that leverages spatiotemporal dynamics for socially compliant navigation. To represent social context based on interactions, we introduce a graph-based auxiliary network that encodes crowd states by attending to pedestrians. In addition, we present a navigation module that captures temporal dynamics and mitigates error accumulations by incorporating encoded state predictions and employing a trajectory-level learning objective. Our framework outperforms established data-driven baselines on simulation and a real-world dataset across diverse social metrics.

\end{abstract}

\section{Introduction}

The use of mobile robots for service or delivery tasks has increased significantly in places including warehouses, airports, and hospitals. Therefore, autonomous mobile robots are expected to navigate in human-populated environments while exhibiting socially acceptable behaviors. An important property of social navigation is that it seeks to minimize the disturbance of the pedestrians~\cite{singamaneni2024survey, francis2025principles}. To meet this property, humans' behavioral patterns are integrated, either implicitly or explicitly, into the decision-making process of the robot navigation~\cite{mavrogiannis2023survey}.

An increasing number of studies utilize Learning from Demonstrations (LfD) methods, which enable supervised learning of social navigation from real-world human demonstration data without requiring hand-crafted rules or engineered reward functions, offering high flexibility. In this context, Inverse Reinforcement Learning (IRL) methods aim to extract a reward function from expert demonstrations~\cite{kretzschmar2016socially} and use it to learn a policy by reward optimization. The learned reward functions are capable of capturing complex behaviors; however, since the reward optimization must be repeated multiple times during training, IRL methods struggle with computational efficiency~\cite{ho2016generative}.
A computationally lighter, widely used LfD method is Behavior Cloning (BC), which learns a mapping from states to actions by imitating expert demonstrations. BC is relatively simple to train and deploy since it does not require active exploration in the environment or any reward specification. However, since the agent has only seen the states in the training data, a small shift in the test distribution will cause accumulated errors due to the buildup of single-step decisions, a recognized challenge in sequential decision-making problems such as navigation~\cite{ross2011reduction}.
To address this limitation, Tai et al.~\cite{tai2018socially} utilized a trajectory-level learning objective via Generative Adversarial Imitation Learning, which evaluates entire sequences rather than single timesteps, mitigating error accumulation. Still, its adversarial training is unstable due to the joint optimization of the discriminator and the policy, and it is sample-inefficient because it requires on-policy rollouts at each iteration. 
Similarly, Yildirim and Ugur~\cite{yildirim2022learning} used Conditional Neural Processes~\cite{garnelo2018conditional} with a trajectory-level learning objective and a fixed number of nearest pedestrians for social navigation. 
However, a representation that encodes social interactions to anticipate pedestrian motion is required for more nuanced, socially aware maneuvers.

Beyond the learning paradigm, the performance of the navigation model depends on its ability to 
anticipate pedestrian motion.
Common approaches consider only the current state or state history to plan and generate motions~\cite{singamaneni2024survey}. However, navigating based solely on former observations may limit the foresight to capture pedestrians' intentions. For example, distant people may appear safe, but their paths could lead to a future collision with the robot.
Therefore, incorporating anticipated pedestrian movements into navigation decisions is crucial for executing human-aware preemptive maneuvers. Prior works addressed this issue mostly explicitly, by separating human trajectory prediction and feeding these estimations to a policy~\cite{sathyamoorthy2020densecavoid}. These methods are interpretable, but they are computationally intensive and are prone to accumulated errors in predictions as they require a separate network to predict the future states. To overcome these limitations, an implicit approach without the overhead of a separate prediction mechanism is proposed.

Pedestrians in densely populated environments adapt their movements to their surrounding crowd, which contains both human-human and human-robot interactions. Previous works either used a fixed number of pedestrians or represented the crowd state through independent pairwise human-robot interactions~\cite{chen2019crowd}, which limits the capture of the relational structure within the group of pedestrians. This problem can be addressed by Graph Neural Networks (GNNs), which embed agents through processing a graph-based state of the crowd, where each agent is a node and their interactions are edges. Relational Graph Learning (RGL)~\cite{chen2020relational} used GNNs to learn the crowd structure and showed success over methods that did not integrate human-human interactions. However, RGL learns the crowd representation jointly with its reinforcement learning policy, so the representations are shaped by the reward signal rather than the crowd observation itself. In contrast, we train a novel crowd representation learning method, which we name Graph Feature Autoencoder (GFAE) as a separate module, producing transferable crowd embeddings that capture social interactions independent from downstream tasks.

In this study, we propose a social navigation framework that addresses the limitations of prior methods: (1) \textit{error accumulation from single-step training}, (2) \textit{neglect of anticipated pedestrian movements}, and (3) \textit{limited modeling of social structure}. Our contributions can be listed as follows:
\begin{itemize}
    \item We present a spatiotemporal framework for social robot navigation that integrates learned spatial crowd representation with temporal trajectory learning.

    \item We introduce a navigation module trained with a trajectory-level learning objective to \textit{mitigate error accumulation}. Furthermore, it \textit{incorporates pedestrian movement estimations} into navigation decision-making by jointly predicting them alongside the robot trajectory from a shared latent representation.

    \item We propose a novel method named Graph Feature Autoencoder that encodes \textit{social structure of the crowd} into a fixed-size crowd embedding by leveraging human-human and human-robot interactions.
\end{itemize}
\section{Related Work}

Robot navigation in human-populated environments has been researched for decades. Pioneer works tried to model the pedestrian motion with predefined rules. Social Force Model (SFM)~\cite{helbing1995social} describes pedestrian motion through a system of social forces that attract individuals toward a goal and repel them from others. Similarly, Optimal Reciprocal Collision Avoidance (ORCA)~\cite{van2011reciprocal} guarantees collision-free navigation in a multi-agent environment by assuming agents move cooperatively. However, since these rule-based methods rely on manually defined utility metrics, they struggle with dense, unpredictable crowds, often leading to the 'Freezing Robot Problem' where the robot fails to find a safe local path to move~\cite{mavrogiannis2023survey}.

To overcome the limitations of rule-based approaches, there is a line of work that uses Deep Reinforcement Learning (DRL), a paradigm that aims to approximate a target policy that maximizes a predefined reward function.~\cite{chen2017decentralized, chen2017socially, han2025dr} uses DRL to navigate in multi-agent environments. Since it is difficult to define a reward function to capture social norms, DRL methods need a careful reward-engineering process that requires expert knowledge.
Additionally, training RL agents on real robots is both unsafe, as the exploratory actions risk harming nearby humans, and expensive, as long-term physical interaction with the environment can result in high operational costs. This leads to training in simulated environments, which brings along a sim-to-real gap that limits performance, safety, and robustness.

In contrast to DRL, the Learning from Demonstrations framework enables to learn directly from real-world data and does not suffer from the aforementioned problems.~\cite{kretzschmar2016socially} used Inverse Reinforcement Learning (IRL) to model pedestrian behavior with hand-crafted features and showed reasonable capability, though being computationally inefficient~\cite{ho2016generative}.~\cite{yildirim2022learning} used Conditional Neural Processes for a data-driven navigation framework, but assumed a fixed number of pedestrians, and assigned equal importance to all pedestrians regardless of their movements. Recently,~\cite{song2025vlm} proposed a foundation model-based approach to social navigation. While being robust to different scenarios, utilizing a foundation model requires inference from a centralized server, which is problematic for achieving truly independent on-the-field mobile robots. 

Graph-based state representations that integrate an attention mechanism have become popular to capture social dynamics in related tasks.~\cite{jain2016structural, vemula2018social} used spatiotemporal graph representations for human trajectory prediction.~\cite{chen2020relational, chen2020robot, chen2024socially} combined RL policies with graph representations that model human-populated environments, to learn robot navigation. However, these methods did not utilize future state predictions in their navigation decision-making. Furthermore, these approaches are prone to the general limitations of the reinforcement learning paradigm, as previously mentioned. 
\section{Method}

\subsection{Preliminaries}

\subsubsection{Graph Neural Networks}
GNNs are specialized neural networks that operate on graph-structured data. Formally, a graph is defined as $G= (V, E)$ where $V$ is the set of nodes and $E$ is the set of edges between the nodes. Some examples of graph-structured data are social networks and molecules, both of which are conveniently represented as objects (nodes) and their relations (edges). GNNs employ a graph convolution operation over each node's neighborhood to generate node embeddings that carry relational information of the local graph structure. For a given node $i \in V$ with input features $\mathbf{x}_i$, graph convolution operation~\cite{Kipf2016gcn} updates the node's representation $\mathbf{h}_i$ by aggregating features from its local neighborhood $\mathcal{N}(i)$, including itself:
\begin{equation}
    \mathbf{h}_i = \sigma \left( \sum_{j \in \mathcal{N}(i) \cup \{i\}} c_{ij} \mathbf{W} \mathbf{x}_j \right)
\end{equation}
where $\sigma$ is a non-linear activation function, $\mathbf{W}$ is a learnable weight matrix, and $c_{ij}$ is a normalization coefficient based on graph structure.
Prior work proposed to integrate the attention mechanism into the graph convolution operation by replacing the static normalization coefficients $c_{ij}$ with dynamically computed attention coefficients, as demonstrated by
Graph Transformer Layer~\cite{ijcai2021p214}, which utilizes the multi-head attention mechanism~\cite{vaswani2017attention}.

\subsubsection{Conditional Neural Movement Primitives}

CNMP~\cite{seker2019conditional} is a Learning from Demonstrations framework that extracts complex temporal relations in the robot's sensorimotor space by learning conditional distributions across uniformly sampled observation and target timesteps. During training, CNMP samples a set of observations $\mathcal{O} = \{(t_k, SM(t_k))\}_{k \in \mathcal{T}_{obs}}$ from a demonstration where $\mathcal{T}_{obs}$ is the set of observation timesteps, $t_k$ and $SM(t_k)$ are the time and the sensorimotor value at timestep $k$, respectively. Then, it predicts a conditional probability distribution in the sensorimotor space given a target query time $t_{tar}$ and sampled observations $\mathcal{O}$. The architecture consists of an encoder $\text{Enc}_{\psi}$ that maps the observations to a latent space, a mean aggregation to produce a single latent representation $\mathbf{r}$, and a decoder $\text{Dec}_{\phi}$ that takes $\mathbf{r}$ and a target query time $t_{tar}$ to output Gaussian distribution parameters $\hat{\boldsymbol{\mu}}, \hat{\boldsymbol{\sigma}}^2$. The weights of encoder $\text{Enc}_{\psi}$ and decoder $\text{Dec}_{\phi}$, $\psi$ and $\phi$, respectively, are optimized to minimize the Gaussian negative log-likelihood of the true sensorimotor value at target timestep $SM(t_{tar})$:
\begin{equation}
    \label{eq:cnmpdataflow}
    \hat{\boldsymbol{\mu}}, \hat{\boldsymbol{\sigma}}^2 = \text{Dec}_{\phi} \left( \frac {\sum_{k \in \mathcal{T}_{obs}} \text{Enc}_{\psi}(t_k, SM(t_k))}{|\mathcal{T}_{obs}|}, t_{tar} \right)
\end{equation}
\begin{equation}
    \label{eq:cnmploss}
    \mathcal{L}(\psi, \phi) = - \mathbb{E} \left[ \log \mathcal{N} \left(SM(t_{tar}) ;  \hat{\boldsymbol{\mu}}, \hat{\boldsymbol{\sigma}}^2 \right) \right]
\end{equation}
After training, a trajectory can be generated by giving the current observation as input and querying the model with all future timesteps. 
\subsection{Problem Definition}
\begin{figure*}[!ht]
    \centering
    \includegraphics[width=0.9\linewidth]{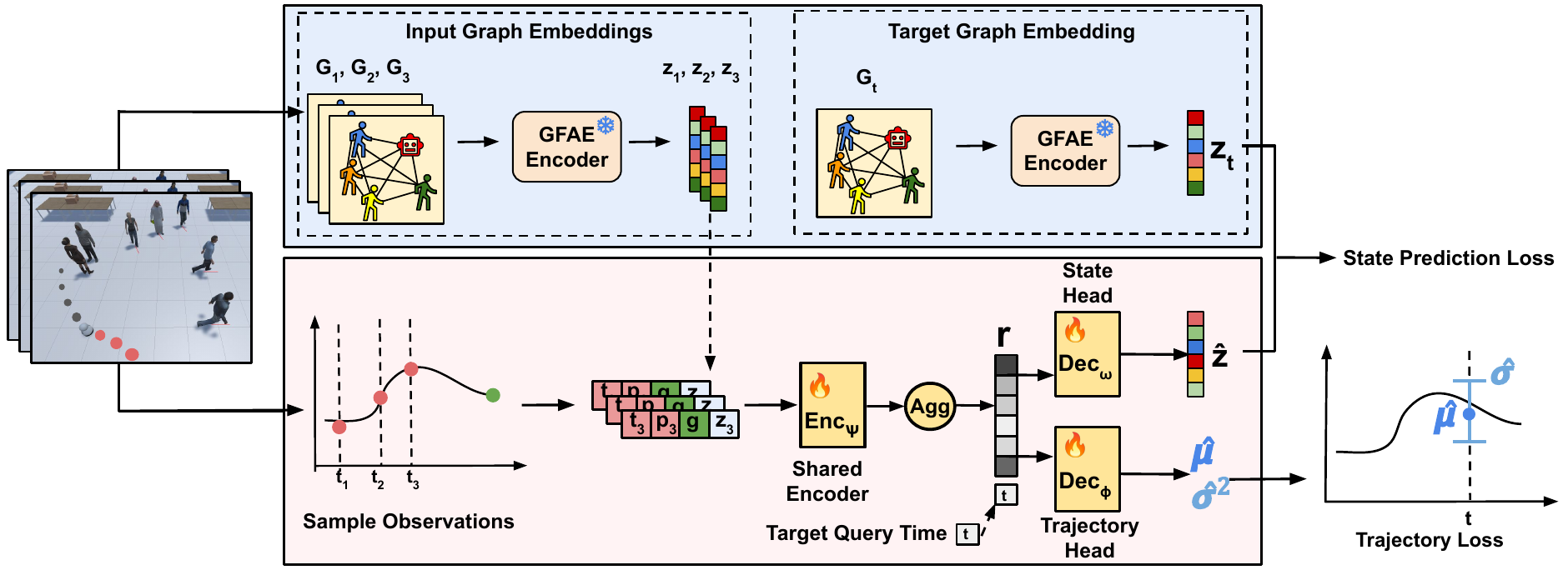} 
    \caption{ Training procedure of the trajectory generator network. First, GFAE is trained to extract graph embeddings to be used as state representations. Then, sampled observations from a demonstration are given to a shared encoder $\text{Enc}_{\psi}$, parametrized by $\psi$, that embeds each observation into a latent vector. An aggregation is applied to the hidden vectors to construct a shared representation $\mathbf{r}$ that will be by both the state head $\text{Dec}^{state}_{\omega}$ and trajectory head $\text{Dec}^{traj}_{\phi}$, parametrized by $\omega$ and $\phi$, respectively. All weights are jointly optimized to minimize the linear combination of the state and trajectory prediction losses.}
    \label{fig:gcnmp}
\end{figure*}
 
    Our objective is to propose a learning-based social navigation framework for autonomous mobile robots navigating in human-populated environments. In our setting, we aim to approximate a function that, given the current robot position, pedestrians' positions, and a goal, generates the socially optimal trajectory enabling the robot to reach its goal. 
    For this, we leverage a dataset $\mathcal{D}$ composed of socially compliant expert demonstrations. Firstly, we define the key terminology that is later used in the explanation of our method.

    A \textbf{trajectory} is an ordered set of timestamped positions, $\tau = \{(t_0, \mathbf{p}_0),\ldots,(t_T, \mathbf{p}_T)\}$, where $t \in \mathbb{R}$ and $\mathbf{p}_t \in \mathbb{R}^2$ denotes time and 2D position, respectively. 
    
    A \textbf{state} is an unordered set containing positions of $N + 1$ \textbf{agents} ($N$ pedestrians and one controllable robot),  $s_t = \{\mathbf{p}^{robot}_t, \mathbf{p}^{ped, 1}_t, \ldots, \mathbf{p}^{ped, N}_t\}$. ${\mathbf{p}}^{robot}_t \in \mathbb{R}^2$ and $\mathbf{p}^{ped, i}_t \in \mathbb{R}^2$ denote the robot's and $i^{th}$ pedestrian's 2D positions, respectively.

    A \textbf{graph-based state} is an undirected complete graph, $G_t = (V_t, E_t)$, where each node in the graph is an agent in the environment (robot or a pedestrian) that contains its position as features, $V_t = \{ v^{robot}_t, v^{ped, 1}_t, \ldots, v^{ped, N}_t\}, E = \{ (v^i, v^j) \mid v^i, v^j \in V, i \neq j\}$.

\subsection{Method Overview}

\algnewcommand{\LineComment}[1]{\State \(\triangleright\) #1}
\begin{algorithm}[!b]
\caption{Training Process}
\label{alg:train}
    \begin{algorithmic}[1]
    \Require Training Dataset $\mathcal{D} =  \{ \tau^i \}_{i=1}^M$ where each trajectory $\tau^i = \{t^i_j, \mathbf{p}^i_j, G^i_j \}_{j=1}^T ; \text{hyperparameter} \; \lambda $
    
    \State \textbf{Stage 1: Train GFAE}
    \For{each mini-batch $\mathcal{B}$ in $\mathcal{D}$}
        \State $\hat{\mathbf{X}}, \mathbf{z} \gets \text{GFAE}_\theta(G_{\mathcal{B}})$
        \State Update $\theta$ w.r.t. $MSE(\mathbf{X}, \hat{\mathbf{X}})$
    \EndFor
    
    \State \textbf{Stage 2: Train CNMP}
    \For{each mini-batch $\mathcal{B}$ in $\mathcal{D}$}
        \LineComment{Sample observations and targets}
        \State $\mathcal{O}, \mathbf{p}_{tar}, \mathbf{z}_{tar} \sim \mathcal{B}$
        \LineComment{Query model and compute loss}
        \State $\mathbf{r} \gets \text{Agg}(\text{Enc}_\psi(\mathcal{O}))$
        \State $\hat{\boldsymbol{\mu}}, \hat{\boldsymbol{\sigma}}^2 \gets \text{Dec}^{traj}_{\phi}(\mathbf{r}, \mathcal{T}_{tar})$; \quad $\hat{\mathbf{z}} \gets \text{Dec}^{state}_{\omega}(\mathbf{r}, \mathcal{T}_{tar})$
        
        \State $\mathcal{L}_{traj} \gets - \log \mathcal{N}(\mathbf{p}_{tar}; \hat{\boldsymbol{\mu}}, \hat{\boldsymbol{\sigma}}^2)$
        \State $\mathcal{L}_{state} \gets \| \mathbf{z}_{tar} - \hat{\mathbf{z}}\|^2$
        \State Update $\psi, \phi, \omega$ w.r.t. $\mathcal{L}_{traj} + \lambda\mathcal{L}_{state}$
    \EndFor
    
    \end{algorithmic}
\end{algorithm}

We introduce a framework that decouples crowd representation learning from navigation policy learning through a two-stage training pipeline (Alg.~\ref{alg:train}). In the first stage, we train the Graph Feature Autoencoder to encode graph-based crowd states into fixed-size crowd embeddings. In the second stage, we train a navigation module extending the CNMP~\cite{seker2019conditional} architecture. This module uses these embeddings to jointly predict the robot's trajectory and future crowd embeddings from a shared latent vector, as shown in Figure~\ref{fig:gcnmp}.

\subsection{Graph Embeddings as State Representations}

\begin{figure}[!t]
    \centering
    \includegraphics[width=\linewidth]{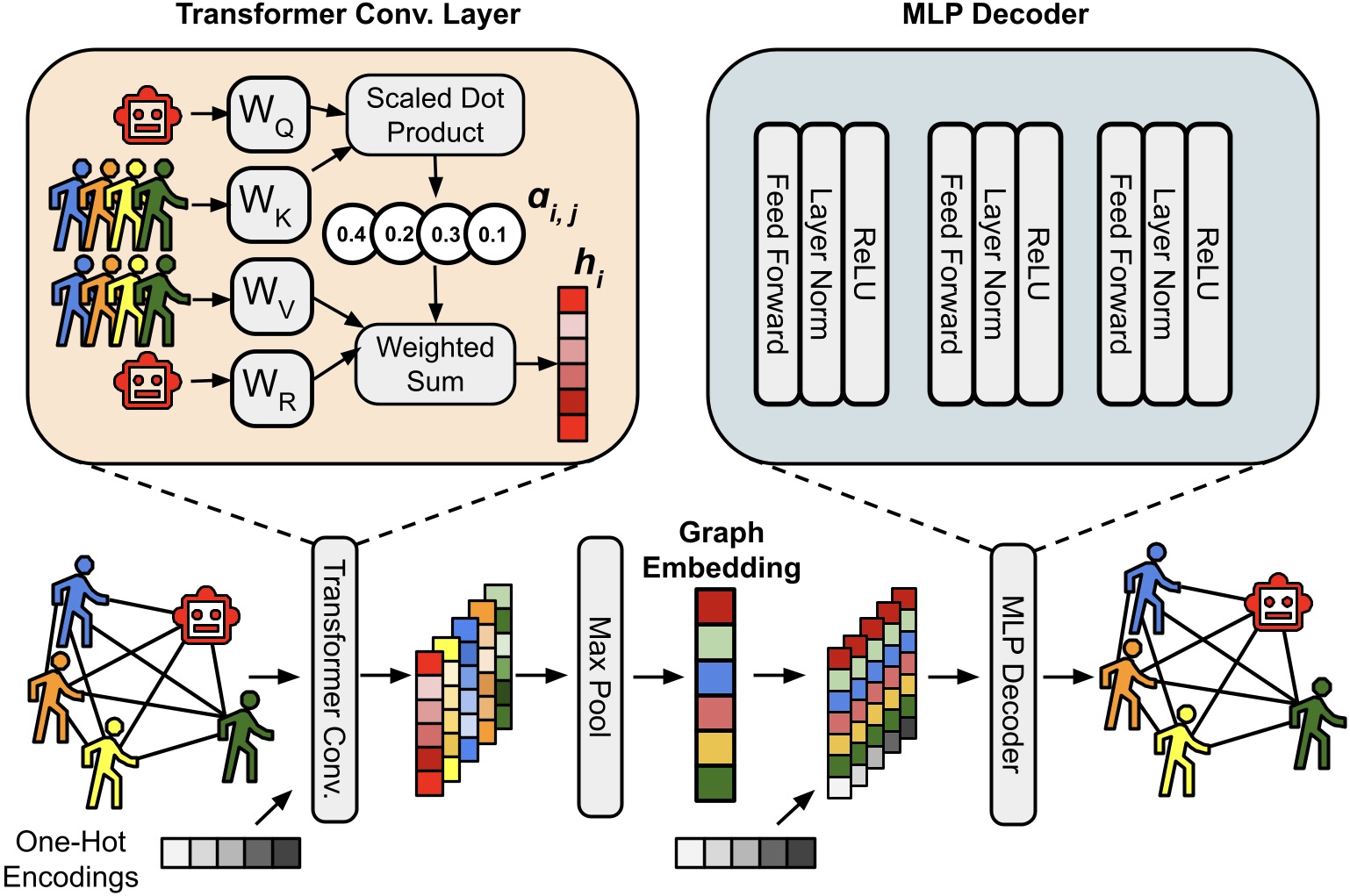} 
    \caption{The architecture of Graph Feature Autoencoder. Firstly, Graph Transformer Layer (GTL)~\cite{ijcai2021p214} takes node features $\mathbf{X}$ and adjacency matrix $\mathbf{A}$ as input, and produces node embeddings $\mathbf{h}$ for each node, by calculating the attention coefficients $\alpha_{ij}$ and integrating them into a weighted sum. GTL contains 4 different weight matrices: $\mathbf{W}_Q$, $\mathbf{W}_K$, $\mathbf{W}_V$, and $\mathbf{W}_R$, which are the query, key, value, and residual weights, respectively. Then, node embeddings are pooled to extract a single graph embedding. This graph embedding is concatenated with each node's one-hot encoding vector and passed through the Multi-Layer Perceptron (MLP) decoder 
    to reconstruct the corresponding node features. Model weights are trained with respect to the mean squared loss between the original $\mathbf{X}$ and reconstructed $\hat{\mathbf{X}}$ node features.}
    \label{fig:gae}
\end{figure}

A socially informative crowd representation captures how pedestrians may influence one another, in addition to their individual conditions. A primary factor that determines these pairwise influences is proximity: closer pedestrians influence each other more than distant ones. Therefore, pairwise relations should be weighted accordingly, rather than treated equally. Moreover, a module that generates such crowd representations should be able to process a varying size of input, to efficiently encode crowds with different numbers of pedestrians. This variable input size makes feedforward neural networks incompatible with this task since they require the crowd to be padded or truncated to a predetermined size.

We propose the Graph Feature Autoencoder (GFAE) to satisfy both requirements.
In the GFAE architecture illustrated in Fig.~\ref{fig:gae}, we leverage GNNs~\cite{Kipf2016gcn} to process crowds with varying numbers of pedestrians, and an attention mechanism~\cite{ijcai2021p214, vaswani2017attention, veličković2018graph} to weight human-human and human-robot interactions according to their influence. The goal of this network is to create a crowd representation to be used as input to the downstream trajectory generator, which is a feedforward network. Therefore, we obtain a fixed-size crowd embedding through global pooling of node embeddings. To ensure this aggregation preserves per-node information, we train a decoder to reconstruct each node's features from the pooled crowd embedding. This feature reconstruction motivates the name Graph Feature Autoencoder.

We now detail the architecture of this module, which consists of an encoder, a pooling mechanism, and a decoder. Graph Transformer Layer (GTL)~\cite{ijcai2021p214} is used as the encoder. GTL uses the multi-head attention mechanism~\cite{vaswani2017attention} to produce $N + 1$ node embeddings $\{\mathbf{h}_i\}_{i=1}^{N+1}$. Given node features $\mathbf{X}$, each attention coefficient $\alpha_{ij}$ is obtained by the following equations:

\begin{equation}
    \beta_{ij} = \frac{(\mathbf{W}_Q \mathbf{x}_i)^\top(\mathbf{W}_K \mathbf{x}_j)}{\sqrt{d}}
\end{equation}
\begin{equation}
    \alpha_{ij} = \frac{\text{exp}(\beta_{ij})}{\sum_{k \in \mathcal{N}(i)} \text{exp}(\beta_{ik})}
\end{equation}
where $\mathbf{W}_Q$ and $\mathbf{W}_K$ represent the query and key weight matrices, $d$ represents the dimension of the hidden layer, $\beta_{ij}$ represents the result of the scaled dot product operation. Normalized attention coefficients $\alpha_{ij}$ are obtained by applying softmax over the neighborhood of $\mathbf{x}_i$. Our motivation is to differentiate strong and weak social interactions by the attention coefficients: intuitively, spatially closer agents are likely to apply stronger social influence on each other. This way, each node embedding is constructed with a sum over neighborhood embeddings that is weighted on the social influence (attention coefficient):
\begin{equation}
    \mathbf{h}_i = \mathbf{W}_R \mathbf{x}_i + 
    \sum_{j \in \mathcal{N}(i)} \alpha_{ij} \mathbf{W}_V \mathbf{x}_j
\end{equation}
where $\mathbf{W}_R$ and $\mathbf{W}_V$ are the residual and value weight matrices. Then, max pooling is applied to node embeddings to produce a fixed-size crowd embedding $\mathbf{z}$. Note that different
pooling mechanisms can be applied in this process, such as mean or attention pooling; we found that max pooling performed best among different common pooling choices. 

Lastly, the decoder should reconstruct each node's features using this single crowd representation $\mathbf{z}$. This poses two challenges. First, the MLP decoder's output dimension is fixed, whereas the size of the reconstructed graph varies with the number of pedestrians. Second, the pooled embedding aggregates all node features into a single vector, so reconstructing a specific node requires an additional index for identifying it. 
To overcome these issues, each node is indexed with a one-hot encoding vector before passing it through the encoder. Then, any node's features can be predicted by passing the graph embedding alongside the corresponding one-hot vector to the MLP decoder.
All node features can be reconstructed this way to produce $\hat{\mathbf{X}}$. GFAE is trained to minimize the mean squared error between the original and reconstructed node features (Alg.~\ref{alg:train}, lines~3--4).
After training, we extract graph embeddings $\mathbf{z}$ and use them as state representations in downstream feedforward networks.

\subsection{Conditional Robot Trajectory Generation}
The next component of our framework is the navigation module, which generates socially compliant robot trajectories based on the learned graph embeddings. While standard Behavior Cloning (BC) suffers from error accumulation under distribution shifts, a trajectory-level objective that encourages learning temporal relations between different states and actions can mitigate this problem~\cite{tai2018socially}.
In this context, we propose a navigation module that builds upon the CNMP framework~\cite{seker2019conditional}, where the sensorimotor value $SM(t)$ is the robot position $\mathbf{p}_t$, and the observation set $\mathcal{O} = \{(t_k, \mathbf{p}_k, \mathbf{z}_k, \mathbf{g}_k)\}_{k \in \mathcal{T}_{obs}}$ includes observation tuples that consist of time, robot position, graph embedding and goal position, respectively. This way, we can utilize fixed-size graph embeddings $\textbf{z}$ that are extracted from GFAE. A compact latent representation $\textbf{r}$ is constructed by averaging encoded observation points (Alg.~\ref{alg:train}, line~11):
\begin{equation}
    \textbf{r} = \frac{\sum_{k \in \mathcal{T}_{obs}}\text{Enc}_{\psi}(t_k, \mathbf{p}_k, \mathbf{z}_k, \mathbf{g}_k)}{|\mathcal{T}_{obs}|}
\end{equation}
The decoder $\text{Dec}^{traj}_\phi$ predicts Normal distribution parameters $\hat{\boldsymbol{\mu}}, \hat{\boldsymbol{\sigma}}^2$ of robot position given this latent representation $\textbf{r}$ and a target query time $t_{tar}$. To incorporate pedestrians' trajectory estimations flexibly, we introduce a \textbf{new state decoder} $\text{Dec}^{state}_\omega$ that predicts the crowd state (graph embedding) $\hat{\textbf{z}}$ given the same latent representation and target query time. Therefore, in a single forward pass, given a target query time $t_{tar}$, both the state and the robot position are jointly predicted from a shared latent vector (Alg.~\ref{alg:train} line 12). Although the heads are separate, since both are using the same latent vector, the shared encoder $\text{Enc}_\psi$ is regularized to produce a representation that holds two different but related pieces of information: where the robot should be and how the pedestrians will move. As stated in the previous section, graph embeddings contain node feature information, in our case pedestrian positions, and therefore, they implicitly carry the information of social dynamics.

Fig.~\ref{fig:gcnmp} illustrates the training process, where both observation and target timesteps are sampled uniformly from a demonstration in the dataset (Alg.~\ref{alg:train} line~9). The figure depicts a target timestep in the future with respect to the observations, providing an intuitive example of jointly learning future states and robot positions. However, this causal structure is not always the case. Since both observation and target timesteps are uniformly sampled, the targets can be arbitrarily placed in the temporal axis; before, after, or in between the observation points. Compared to the BC objective, this setting is relatively hard since it forces the latent representations to encode information that covers the joint space of crowd state and the robot's trajectory, across the temporal axis of demonstrations. In this way, learning is conducted on the trajectory-level, and the navigation module alleviates the problems of BC setting due to its short-horizon learning objective. The network parameters are optimized with the linear combination of individual prediction losses (Alg.~\ref{alg:train} line~15):
\begin{equation}
    \mathcal{L}_{total} = \mathcal{L}_{traj} + \lambda \mathcal{L}_{state}
\end{equation}
where $\lambda$ is a hyperparameter, $\mathcal{L}_{traj}$ is the Gaussian negative log-likelihood loss for trajectory prediction (Alg.~\ref{alg:train} line~13), and $\mathcal{L}_{state} $ is the mean squared loss for state prediction (Alg.~\ref{alg:train} line~14).
At inference time, a graph-based state of the crowd is constructed and given as input to GFAE to extract a graph embedding $\mathbf{z}$. Then, the current observation tuple $(t, \mathbf{p}, \mathbf{z}, \mathbf{g})$ that includes the current robot position $\mathbf{p}$, a goal position $\mathbf{g}$ is fed to the network. A trajectory is generated by querying the trajectory decoder with the latent representation $\mathbf{r}$ and all future timesteps. 
\section{Experiments}

We conduct experiments on simulation and real-world datasets to verify our three main claims:
\begin{itemize}
    \item Our method that learns temporal relations between the sensorimotor and the crowd embedding space exhibits better social performance than common reactive methods.
    \item Using learned graph embeddings improves the performance
    of navigation methods, compared to using concatenated vectors of pedestrian positions directly as state representations.
    \item Our method is able to learn from in-the-wild real-world data and enable a simulated robot to navigate in a socially compliant manner.
\end{itemize}

\subsection{Simulation Experiments}

\begin{table*}[!ht]
    \centering
    \caption{Comparison of navigation baselines (mean $\pm$ std.) over 5 runs. The results are reported across six metrics that are defined in Section~\ref{sec:evaluation}. Bold values indicate the best results and those not significantly different (paired t-test, $p>0.05$).
    }
    \label{tab:nav_results_grid}
    \begin{tabular}{lcccccc}
        \toprule
        \textbf{Method} & \textbf{IZIC $\downarrow$} & \textbf{PZIC $\downarrow$} & \textbf{MDTP (m) $\uparrow$} & \textbf{Collisions $\downarrow$} & \textbf{Navigation Time (s) $\downarrow$} & \textbf{Success Rate (\%) $\uparrow$} \\
        \midrule
        BC  & 0.26 $\pm$ 0.07 & 1.38 $\pm$ 0.19 & 0.97 $\pm$ 0.07 & 0.12 $\pm$ 0.10 & \textbf{31.27 $\pm$ 0.56} & 
        88.12 $\pm$ 8.09 \\
        Diffusion Policy  & 0.20 $\pm$ 0.04 & 1.31 $\pm$ 0.15 & 1.04 $\pm$ 0.03 & 0.07 $\pm$ 0.06 & 32.55 $\pm$ 0.68 & 
        90.62 $\pm$ 3.83 \\
        GE+BC  & \textbf{0.15 $\pm$ 0.04} & 1.38 $\pm$ 0.25 & 1.08 $\pm$ 0.07 & \textbf{0.03} $\pm$ \textbf{0.03} & \textbf{31.09 $\pm$ 1.06} & 
        \textbf{97.50 $\pm$ 2.61} \\
        GE+CNMP (ours)  & \textbf{0.09} $\pm$ \textbf{0.04} & \textbf{1.11 $\pm$ 0.10} & \textbf{1.17} $\pm$ \textbf{0.05} & \textbf{0.03} $\pm$ \textbf{0.04} & 32.12 $\pm$ 0.26 & \textbf{98.12 $\pm$ 2.80} \\
        \bottomrule
    \end{tabular}
\end{table*}

\subsubsection{Environmental Setup}
All simulation experiments are conducted in a warehouse environment inside the SEAN 2.0 simulation~\cite{tsoi2022sean2}, where all pedestrians traverse their path using the Social Force Model~\cite{helbing1995social}. A differential-drive mobile robot is controlled by the human teleoperator in the data collection process, and the learning-based methods in the evaluation process. Throughout the experiments, we used the same low-level controller that generates velocity commands to follow the output trajectories of the compared methods.

\begin{figure}[!t]
    \centering
    \includegraphics[width=\linewidth]{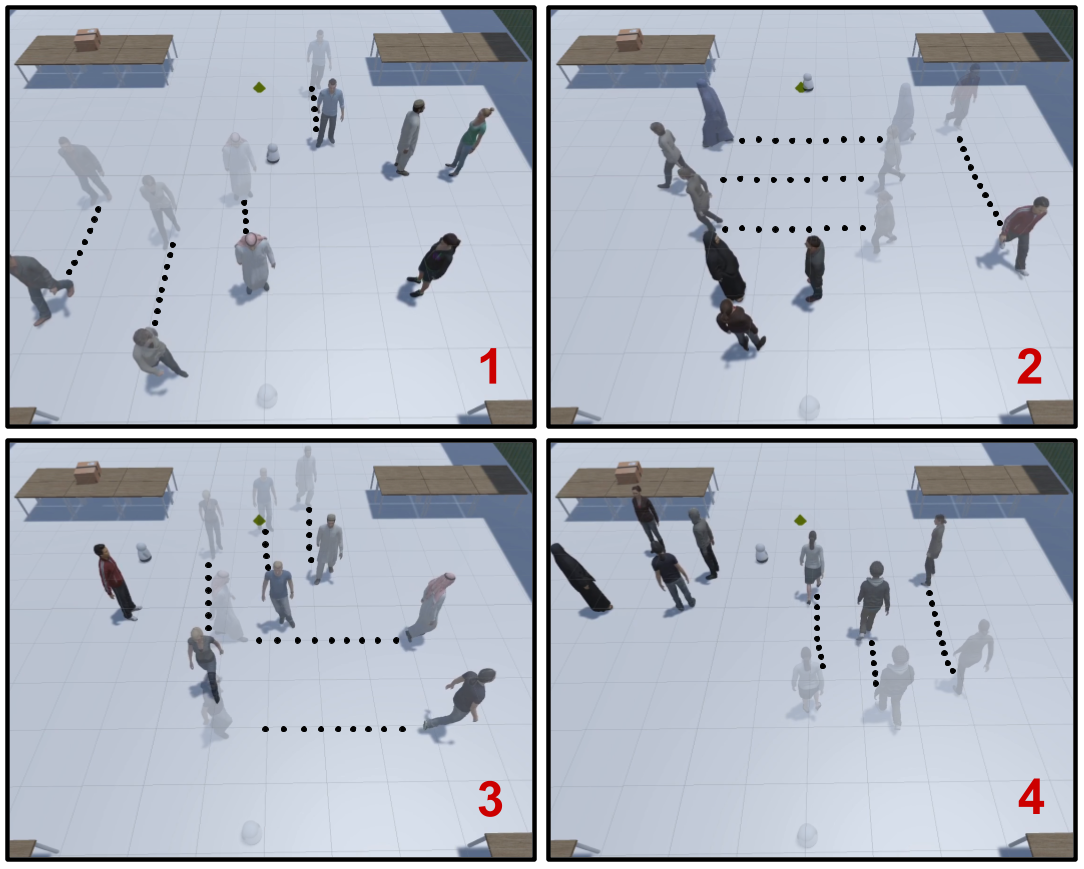} 
    \caption{The illustration of four different social scenarios that we constructed in the SEAN 2.0 warehouse environment. In scenario 1 (top left), the robot needs to pass the upcoming pedestrians without disturbing the standing pedestrians at right. In scenario 2 (top right), the robot should first pass the conversational group and then avoid the walking pedestrians in different ways. In scenario 3 (bottom left), the robot should both cross and pass different pedestrian groups, while avoiding the standing pedestrian at left. In scenario 4 (bottom right), the robot should overtake the pedestrians in front of it, without disturbing the conversational group at left.}
    \label{fig:scenarios}
\end{figure}

\subsubsection{Data Collection}
We collected \textbf{160 demonstrations} by human teleoperation in SEAN 2.0, where each demonstration contains the robot trajectory and all pedestrian positions at each timestep. We included 4 different social scenarios that are based on different human-robot interactions~\cite{francis2025principles}. Each scenario contains one or more of the following interactions: 

\begin{itemize}
    \item Pass walking pedestrian(s),
    \item Overtake walking pedestrian(s),
    \item Cross walking pedestrian(s),
    \item Pass conversational pedestrian(s),
\end{itemize}
 
Furthermore, each scenario contains sub-scenarios where the number of pedestrians varies uniformly between 4 and 7. 
For each sub-scenario, we collected 10 demonstrations, resulting in a total of 160 demonstrations, where each pedestrian's start and goal positions are sampled from a Gaussian distribution with $\sigma = 0.2$ meters. We separated the collected data into a train-validation-test split in a structural way to ensure uniformity: The split is conducted on the sub-scenario level, meaning that for each sub-scenario, the 10 demonstrations are divided into train-validation-test sets. In this way, the splits contain each sub-scenario, but different configurations that are caused by the standard deviation $\sigma$ of the pedestrians' start and goal positions.

\subsubsection{Evaluation}
\label{sec:evaluation}
To evaluate the social performance of the robot trajectory quantitatively, we used metrics proposed by Hall~\cite{hall1966hidden}. The detailed explanation of each metric is as follows:
\begin{itemize}
    \item \textbf{Intimate Zone Intrusion Count (IZIC)} is the number of times the robot passed within 0.45 m of a pedestrian.
    \item \textbf{Personal Zone Intrusion Count (PZIC)} is the number of times the robot passed within 1.20 m of a pedestrian.
    \item \textbf{Minimum Distance To Pedestrians (MDTP)} is the closest distance the robot had to a pedestrian during the execution of a trajectory.
    \item \textbf{Collisions} is the number of times the robot collided with a pedestrian during its path.
    \item \textbf{Navigation Time} is the total time taken by the robot during the execution of a trajectory from start to end.
    \item \textbf{Success Rate} is the robot's success percentage of reaching the given goal position within a small error margin of 0.75 m.
\end{itemize}
We trained the following baselines until convergence and compared their performance on the given metrics.

\begin{figure}[!b]
    \centering
    \includegraphics[width=\linewidth]{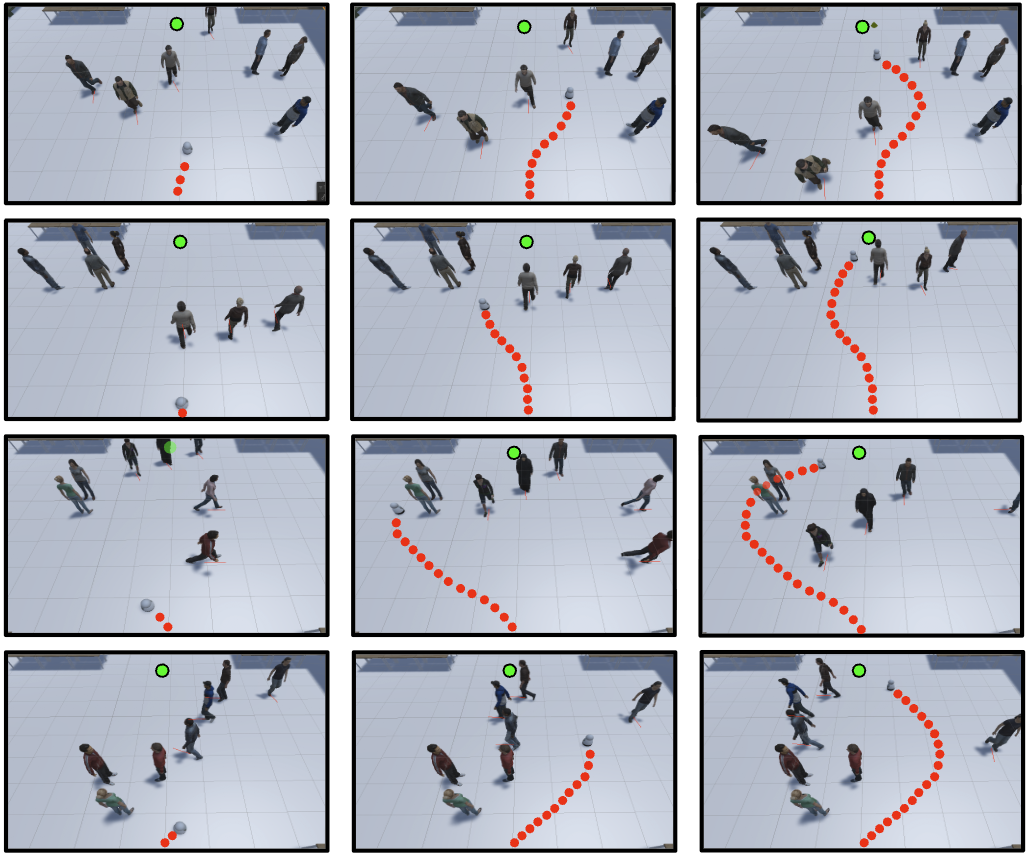}
    \caption{Qualitative snapshots from executed GE+CNMP trajectories on different social scenarios. Each row indicates a unique scenario, where time progresses from left to right. }
    \label{fig:demos}
\end{figure}

\begin{itemize}
    \item \textbf{Behavior Cloning (BC)} policy has the objective of copying the action $\mathbf{a}$ taken in a given state $s$ from the data.  We trained a BC policy with a five-layer multi-layer perceptron. A state is a fixed-size input vector that is the concatenation of the 2D positions of the robot and all pedestrians at that time. We set the size of this vector to $2 \times (N_{max}+ 1)$ and mask out any absent pedestrians, where $N_{max}$ is the maximum number of pedestrians in the demonstrations. Given this state vector, the network predicts the action defined as the difference between consecutive robot position vectors $ \mathbf{p}_{t+1} - \mathbf{p}_{t}$. 
    
    \item \textbf{Diffusion Policy}~\cite{chi2024diffusionpolicy} is an imitation learning algorithm like BC, and policy representation is a conditional diffusion process. During training, we used a Conditional U-net architecture with the same state action formulations as BC, and the original hyperparameters used in~\cite{chi2024diffusionpolicy}. To enable real-time inference, we employed a DDIM Scheduler~\cite{song2021denoising} to accelerate sampling, reducing the denoising steps to 10.
    
    \item \textbf{Graph Embeddings + Behavior Cloning (GE+BC)} is the BC policy where the state representation is the concatenation of the learned graph embedding and the current robot position $ \mathbf{p}_{t}$. The same BC model architecture and action formulation are used during training. In this baseline, our objective is to highlight the benefit of using learned graph-level representations.
    
    \item \textbf{Graph Embeddings + CNMP (GE+CNMP)} is the method we propose, combining the benefits of learned graph embeddings of GFAE and the temporal modeling of CNMP. For the GFAE architecture, we employed a single Graph Transformer Layer as the encoder and a 3-layer MLP as the decoder. For the CNMP architecture, we used 5-layer MLPs for the shared encoder and the separate decoder heads. We set the hyperparameter $\lambda = 1$. 

\end{itemize}

\subsubsection{Results}
The \textbf{social navigation performance} of the learning-based algorithms is provided in Table~\ref{tab:nav_results_grid}. Our method that learns temporal relations performs significantly better than other methods on social and task-wise metrics. This can be explained by two factors: learning temporal connections that enable preemptive movements (1) and utilizing graph embeddings that capture high-level social scenarios (2), both of which enhance socially compliant behavior. Firstly, despite its nearly 80x larger parameter count (67M vs 800K), with a highly expressive policy representation, Diffusion Policy fails to outperform our method.
We attribute this to the lack of a trajectory-level objective that encourages exploiting temporal connections. In addition, the limited demonstration data size (160) may have amplified the performance gap, as diffusion models tend to require more data to generalize. Secondly, GE+BC performs significantly better than BC on most of the metrics, verifying our claim quantitatively that using learned graph embeddings compared to raw states boosts performance on social navigation tasks. Example snapshots from the executed trajectories of our method are provided in Fig.~\ref{fig:demos}, demonstrating that it is capable of exhibiting social maneuvers across different scenarios and numbers of pedestrians.

In addition, we conducted an \textbf{ablation study} on the \textbf{graph convolutional layer} and \textbf{global pooling} choices in the Graph Feature Autoencoder architecture. We included GCNConv layer~\cite{Kipf2016gcn}, GATv2Conv layer~\cite{brody2022how}, and TransformerConv layer~\cite{ijcai2021p214}; mean, max, and attention pooling for the layer and global pooling choices, respectively. The results in Table~\ref{tab:ablation} demonstrate that TransformerConv with max pooling mechanism achieves the lowest reconstruction loss, and the choice of graph layer matters more than the pooling mechanism.

\begin{figure}[!b]
    \centering
    \includegraphics[width=\linewidth]{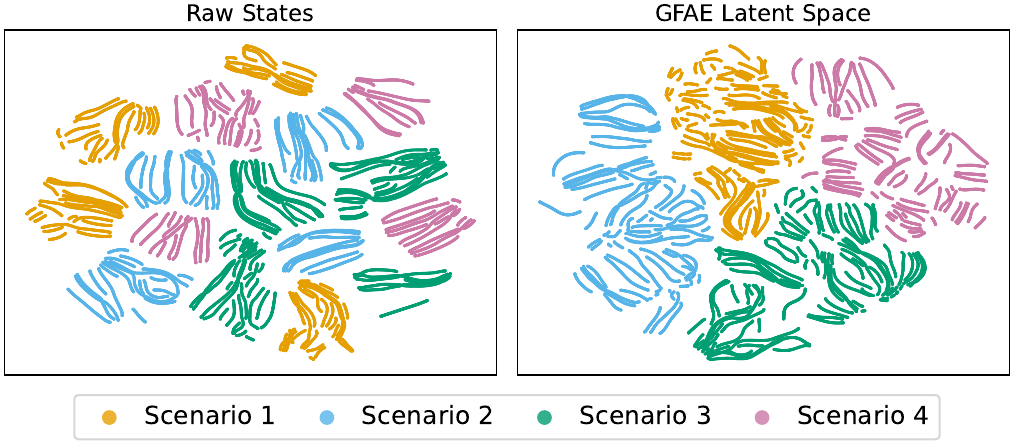}
    \caption{t-SNE visualizations of the simulation dataset where each color represents a different social scenario. The graph embedding space of GFAE~(right) is visually separated by task-driven social scenarios, whereas the raw state space (left) is not.}
    \label{fig:tsne}
\end{figure}

Lastly, we investigate whether the learned graph embeddings effectively encode high-level social scenarios existing within the demonstrations.
Fig.~\ref{fig:tsne} demonstrates the t-SNE~\cite{JMLR:v9:vandermaaten08a} visualizations of the raw state space (left) and the graph embedding state space (right). States from the same social scenarios are clustered together in the graph embedding space, but not in the raw state space. Since we expect agents to move similarly within high-level social scenarios, states clustered by similar scenarios are preferable for learning algorithms. 

\begin{table}[!t]
    \centering
    \caption{Comparison of GFAE Reconstruction Loss (mean $\pm$ std.) across different graph layers and pooling mechanisms over 5 runs. Bold values indicate the best results and those not significantly different (paired t-test, $p>0.05$).
    ($\downarrow$)}
    \label{tab:ablation}
    \begin{tabular}{lccc}
        \toprule
        & \multicolumn{3}{c}{\textbf{Global Pooling Mechanism}} \\
        \cmidrule(lr){2-4}
        \textbf{Graph Layer} & \textbf{Mean} & \textbf{Max} & \textbf{Attention} \\
        \midrule
        GCNConv          & $0.489 \pm 0.069$ & $0.491 \pm 0.070$ & $0.480 \pm 0.067$ \\
        GATv2Conv        & $0.041 \pm 0.009$ & $0.037 \pm 0.007$ & $0.043 \pm 0.009$ \\
        TransformerConv  & \textbf{0.016 $\pm$ 0.006} & \textbf{0.010 $\pm$ 0.002} & \textbf{0.015 $\pm$ 0.007} \\
        \bottomrule
    \end{tabular}
\end{table}

\subsection{Real-World Dataset Experiments}

\subsubsection{Experimental Setup}
We conducted experiments using SCAND~\cite{karnan2022scand} real-world dataset, to verify that our method can learn from noisy, real-world sensor and demonstration data. We utilized the DR-SPAAM~\cite{Jia2020DRSPAAM} detector to extract human positions from 2D LiDAR sensor data. We only used the data from the outdoor environments, where no significant obstacles are present other than pedestrians, to focus on the social aspect of navigation. We divided the data into a train-validation-test split and evaluated the baselines on the test set.
\begin{figure}[!b]
    \centering
    \includegraphics[width=0.8\linewidth]{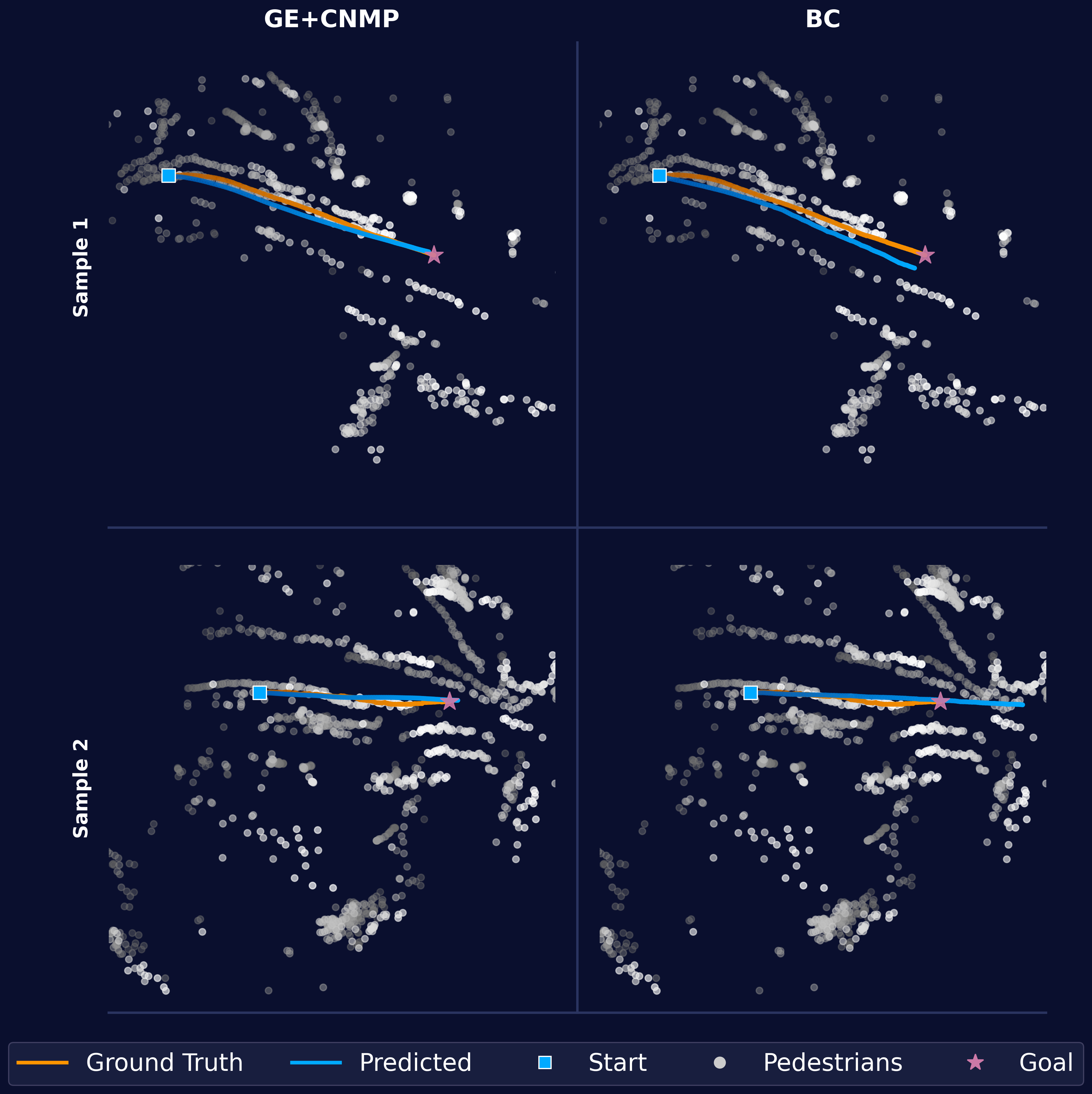}
    \caption{Qualitative plots from SCAND experiments where a row is a sample demonstration and a column is a baseline. White circles are pedestrians, the orange line is the original demonstration, and the blue line is the baseline prediction. A single plot shows the overall execution, where the time progress is represented as a shift from dark to light colors. In Sample 1, BC slightly deviates from the goal towards a pedestrian. In Sample 2, BC overshoots the goal, resulting in a collision with pedestrians. }
    \label{fig:scand_plots}
\end{figure}

\subsubsection{Evaluation}
We evaluated methods on the test set with the recorded pedestrian trajectories from the dataset.
BC and GE+CNMP methods are compared on metrics from the simulation experiments, which are explained in detail under Section~\ref{sec:evaluation}. Each demonstration contains 5 seconds of robot and extracted human positions, recorded at 20 Hz. We also refer to the final robot position of a trajectory as the goal, and condition both networks to reach that destination. We included the nearest 10 humans that are also within 10 meters of the robot into the crowd state.

\subsubsection{Results}

Results in Table~\ref{tab:scand} show that GE+CNMP performs significantly better than BC, 
achieving near social performance to the demonstration data. We observe that BC has a relatively poor success rate, where success is defined as finishing the test within 0.75 meters of the original goal. An example failure is visualized in the bottom right corner of Fig.~\ref{fig:scand_plots}, where BC overshoots the target and collides with the pedestrians. 

\begin{table}[!t]
    \centering
    \caption{Comparison of navigation baselines (mean $\pm$ std.) over 5 runs. Bold values indicate the best performing model, which shows significant improvement (paired t-test, $p<0.05$).}
    \label{tab:scand}
    \begin{tabular}{lccc}
        \toprule
        \textbf{Method} & \textbf{MDTP (m) $\uparrow$} & \textbf{Collisions $\downarrow$} & \textbf{Success Rate (\%) $\uparrow$} \\
        \midrule
        BC & 1.10 $\pm$ 0.08 & 0.95 $\pm$ 0.47 & 53.33 $\pm$ 15.90 \\
        GE+CNMP & \textbf{1.28 $\pm$ 0.05} & \textbf{0.04 $\pm$ 0.08} & \textbf{98.33 $\pm$ 3.33} \\
        \midrule
        Ground Truth & 1.39 $\pm$ 0.08 & 0 $\pm$ 0 & 100 $\pm$ 0 \\
        \bottomrule
    \end{tabular}
\end{table}

\section{Conclusion}

In this work, we propose a social navigation framework that is able to exhibit socially compliant maneuvers by learning from demonstrations, in both simulated and real-world settings. Firstly, we propose a novel auxiliary Graph Feature Autoencoder network that encodes the crowd state into a latent representation by modeling the social interactions with graph neural networks and the attention mechanism. Then, we use these latent vectors as state representations and feed them to a network that jointly predicts future states (these representations) and the robot's navigation trajectory from a shared latent representation, allowing future state estimations to affect navigation decisions in a flexible way. We demonstrated both qualitatively and quantitatively that (1) learned graph embeddings improved performance over default state representations, and (2) our method that incorporated future estimations outperformed other baselines, in both simulated and real-world data.

There are several limitations of our framework. Firstly, the graph embeddings are constructed via pooling over individual node representations, and the scalability of this aggregation operation to highly crowded settings should be verified.
Secondly, we did not integrate any collision avoidance guarantees and constructed the experiments in environments where no static obstacles are present to focus on social maneuvers. In the future, we plan to integrate our framework into a navigation stack, where low-level controllers can ensure collision avoidance, while our method focuses on generating socially compliant trajectories. 

\bibliographystyle{IEEEtran}
\bibliography{ref}
\end{document}